# AI-based artistic representation of emotions from EEG signals: a discussion on fairness, inclusion, and aesthetics

Piera Riccio
ELLIS Unit Alicante Foundation
Oslo Metropolitan University
Campus de San Vicente s/n, 03690
San Vicente del Raspeig, Alicante
Spain
piera@ellisalicante.org

Kristin Bergaust
Oslo Metropolitan University
Postboks 4, St.Olavs plass
0130 Oslo
Norway
kribe@oslomet.no

Boel Christensen-Scheel
Oslo National Academy of the Arts
Fossveien 24
0551 Oslo
Norway
boelchri@khio.no

Juan-Carlos De Martin
Politecnico di Torino
Corso Duca degli Abruzzi, 24
10129, Torino
Italy
demartin@polito.it

Maria A. Zuluaga
EURECOM
450 route des Chappes
06410 Biot, Sophia Antipolis
France
zuluaga@eurecom.fr

Stefano Nichele
Oslo Metropolitan University
Simula Metropolitan Center
for Digital Engineering
Pilestredet 52, 0167 Oslo
Norway
stenic@oslomet.no

**While Artificial Intelligence (AI) technologies are being progressively developed, artists and researchers are investigating their role in artistic practices. In this work, we present an AI-based Brain-Computer Interface (BCI) in which humans and machines interact to express feelings artistically. This system and its production of images give opportunities to reflect on the complexities and range of human emotions and their expressions. In this discussion, we seek to understand the dynamics of this interaction to reach better co-existence in fairness, inclusion, and aesthetics.**

*Brain-Computer Interface. Generative Art. Algorithmic Fairness. Art therapy. Disability Aesthetics.*

## 1. INTRODUCTION

Artificial Intelligence (AI) technologies are having an enormous and complex impact on our cultural production and collective creativity (Manovich, 2017). In this paper, we provide a concrete AI system that generates art pieces representing human emotions detected from electroencephalographic (EEG) signals.

Considerations related to the emotional sphere are naturally restricted when analysing technical and scientific inventions (Picard, 1995). In the same way, we rarely conceive artistic content generated by AI as a representation of ideas or emotions. In our system, humans become providers of emotions, transmitting information that an AI recognizes, encodes, and expresses through a painting. As a result, we explore the grey area between what is human and what is artificial, suggesting the concept of AI as a potential creative extension of humanity.

This work was realized in an Engineering context, aiming to contribute to the technical research on Brain-Computer Interfaces (BCIs) and generative models. The research activity was enriched with constant access to a more humanistic environment in the context of the FeLT Project[1] at Oslo Metropolitan University (Bergaust and Nichele, 2019) emphasising the understanding of philosophical and cultural implications of the proposed system. However, this research is not intended as production of an artistic concept or paradigm. Further developments could expand the work in artistic directions by adding conceptual layers or directing and fine-tuning the system.

---

[1] https://feltproject.no/

In Section 2, we briefly present our pipeline; in Section 3, we analyse its limitations in terms of algorithmic fairness, suggesting future work ideas to overcome them. In Section 4, we provide and comment on some of the obtained paintings. In Section 5, we explore the potentialities of this system in the fields of art therapy and disability aesthetics. Finally, in Section 6, we frame our work in a theoretical discussion, relating it to the existing literature.

## 2. THE GENERATIVE PROCESS

We conceive this research activity as part of the collective effort in understanding how AI is shaping the production of emotionally charged visual images. As such, we have developed our considerations starting from the questions raised by McCormack et al. on generative computer art (McCormack et al., 2014). On this subject, we underline that our generative process relies on the abstraction of the natural physiological response of our brain in the presence of emotional experiences.

Developing an AI system that can deal with emotions requires a challenging mathematical formalization of these vague and subjective phenomena. Historically, psychologists have adopted two different paradigms to characterize emotions: a continuous and a discrete one. The continuous one visualizes an emotion as a point in a multi-dimensional space, with each dimension representing a distinctive characteristic (Russell, 1980). The discrete paradigm, adopted in this work, utilizes a finite set of basic emotions that, when combined, can express more complex feelings (Ekman, 1992).

Our system relies on state-of-the-art deep learning architectures, and it requires two different datasets (one for EEG signals and one for paintings) sharing the same emotional labels. We provide a conceptual pipeline in Figure 1, in which we see the role of both datasets as inputs to two different blocks.

The first block of the pipeline (**EEG emotion encoding**) relies on a Regularized Graph Neural Network (Zhong, Wang and Miao, 2020) dedicated to processing the emotion in the EEG signals. This block produces latent vectors synthesising emotions in EEG signals. We keep the dimension of these vectors high, preserving more information than the single predicted emotion in each signal. This design choice is motivated by an interest in dealing with the influence between different emotions as, in many cases, a single word may not be enough to describe a feeling.

The **image generation block** relies on a generative adversarial network, StyleGAN2ADA (Karras et al., 2020), trained on the dataset of paintings. This block receives the EEG latent vector produced by the **EEG emotion encoding block** and a random vector. The introduction of randomness in generative processes is necessary to simulate the variability and unpredictability of a creative process (McCormack et al., 2014).

For more details on the technical implementation of this system, we refer the readers to (Riccio, 2021).

### 2.1 Related works

To the best of our knowledge, no related work fulfils such a translation of emotional states into paintings. In (Salevati and DiPaola, 2015) and (Colton, Valstar and Pantic, 2008), the authors propose systems to create expressive self-portraits of people. However, these systems have evident limitations, and the users have control over the emotion they want to translate on their portrait (in one case, they select it; in the other, emotions are detected from facial expressions, which can be faked easily). In both works, predefined styles are simply applied to existing portraits. In the context of emotional painting generation from EEGs, we mention (Ekster, 2018) and (Random Quark, 2017). In these cases, the paintings represent emotions through simple lines, predefined shapes, colours, fractals, or bird swarms, causing a rather low inter-painting variability.

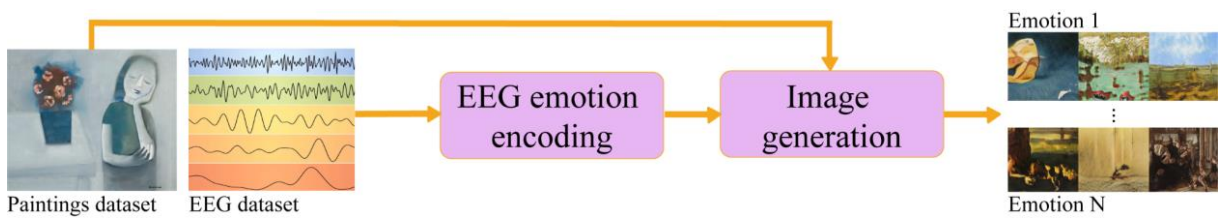

**Figure 1:** Synthetic scheme of the pipeline, from inputs (on the left) to outputs (on the right). EEG image souce: GOQii blog. Painting in the picture: "Room at Twilight" by Charles Blackman (1963).

## 3. ALGORITHMIC FAIRNESS: A CRITICAL PERSPECTIVE

Given the technical and humanistic intentions of this work, we propose a critical perspective on some limitations. To guide this discussion, we introduce algorithmic biases as systematic deviations from a fair representation of reality, sometimes having consequences on specific groups of people (Kordzadeh and Ghasemaghaei, 2021; Ogolla and Gupta, 2018). The presence of algorithmic biases in the **emotion encoding block** of our pipeline could make the interaction with this software rather difficult for some people. At the same time, biases in the **image generation block** could cause an evident under-representation of different cultures in the resulting paintings. Therefore, we consider the analysis of such biases a fundamental step to lean towards a fair system.

### 3.1 Algorithmic biases in EEG emotion encoding

Emotional expression in humans is naturally influenced by cultural identity, to the point that it influences the emotional experience itself (Immordino-Yang, Yang and Damasio, 2016) and causes unwanted social norm violations (Hareli, Kafetsios and Hess, 2015). Studies have shown that the main factors that cause differences in emotional expression are native languages (Wierzbicka, 1999), gender, social contexts, and social position (Drag and Shaw, 1967). In the case of emotion recognition, algorithmic biases can be evident when the recognition is based on facial expressions, word choices in speech, voice tone and other cultural-dependent data modalities (Howard, Zhang and Horvitz, 2017). On the contrary, we do not have information on whether these biases exist in EEG-based emotion recognition. Since EEGs represent an inner physiological reaction, it can be assumed that any existing bias would be less strong. This consideration has led to the choice of recognizing emotions from EEG signals.

We have performed two kinds of experiments: one using the SEED-IV EEG dataset (Zheng et al., 2018) and one performing recordings with an easy-to-use and off-the-shelf device, the OpenBCI headband kit with eight dry-comb electrodes, paired with the Cyton board. The SEED-IV presents high-quality signals, recorded on a homogeneous set of people. The choice to record signals with an off-the-shelf device derives from the desire to create an accessible system, potentially trainable on a diverse set of subjects.

Recording EEG signals ex-novo requires understanding how to elicit emotions in laboratory settings. From an extensive literature review on this topic, we have observed that the most popular choice for emotion stimulation is to utilize cultural contents, like videos (Schaefer et al., 2010; Maffei and Angrilli, 2019), images (Lang, Bradley and Cuthbert, 1997), sounds (Yang et al., 2018). After our experiments, we claim that these contents can considerably produce noise in the elicitation, and the research in the field becomes slower, as stimuli are hard to re-utilize in different cultural contexts. For example, the authors of SEED-IV have publicly shared the emotion-eliciting videos (entirely in Chinese) that they had utilized in their experiments. Despite the subtitles, in our post-experiment interviews we have assessed that these videos are hard to follow for non-Chinese speakers, and we suggest that the research in this area would benefit from further psychological investigations on the methodologies for emotion elicitation.

### 3.2 Algorithmic biases in image generation

The dataset on which generative adversarial networks (GANs) are trained creates a latent space containing the possible features of the generated elements (Goodfellow et al., 2014). Algorithmic biases in content generation can arise unintentionally, creating misalignments between the distributions of the features of the training data versus those of the generated data (Salminen, Jung and Jansen, 2019). In this context, art practices represent a significant opportunity to explore and understand such biases (Booth et al., 2021).

We have trained StyleGAN2ADA on the WikiArt Emotions dataset (Mohammad and Kiritchenko, 2018), a public collection of heterogeneous paintings experimentally labelled with conveyed emotions. Given that these artworks are mainly conceived by western artists, the generated images will inevitably follow the aesthetic norms of western arts, causing a technical limitation. Utilizing a broader dataset (in terms of historical periods and geographical origin) could lead to the generation of paintings that represent emotions relying on cultural-independent and period-independent features. While we can conceive an AI system capable of recognizing these features and reproducing them, such a generalization power is hard to imagine for an individual human artist.

## 4. RESULTS

In this section, we provide a selection of obtained paintings from different experiments at different resolutions. We comment on some of their characteristics, when grouped according to the emotional class of their respective EEG. We trained the pipeline on the SEED-IV (EEG dataset) and WikiArt Emotions (paintings dataset), with labels in four classes: anger, sadness, fear, and happiness.

With image resolution 512x512 pixels, the training took approximately ten days, utilizing the GPU Tesla V100-SXM2-16GB. The training time becomes much longer with less powerful GPUs, and shorter at lower resolutions. The resulting images were upscaled using a super-resolution technique (Sun and Chen, 2020). Although acknowledging that connecting images to emotions is a cultural and learned experience, we mention some features in the paintings.

In "sadness" (Figure 2), it is possible to observe a prevalence of blue shades and cold colours. Regardless of the colour palette, most of the paintings seem to depict characters, scenes, or objects that transmit a sense of loneliness and abandonment. Among them, it seems possible to discern a nomad amid a storm, two figures hugging and crying, a broken vase in the shape of a human head, or an empty room with empty shelves. In "anger" (Figure 3), we observe warmer and darker shades becoming prevalent in the paintings, and the depicted characters are mostly demonic figures or skulls. The atmospheres have a strong imaginative and abstract component.

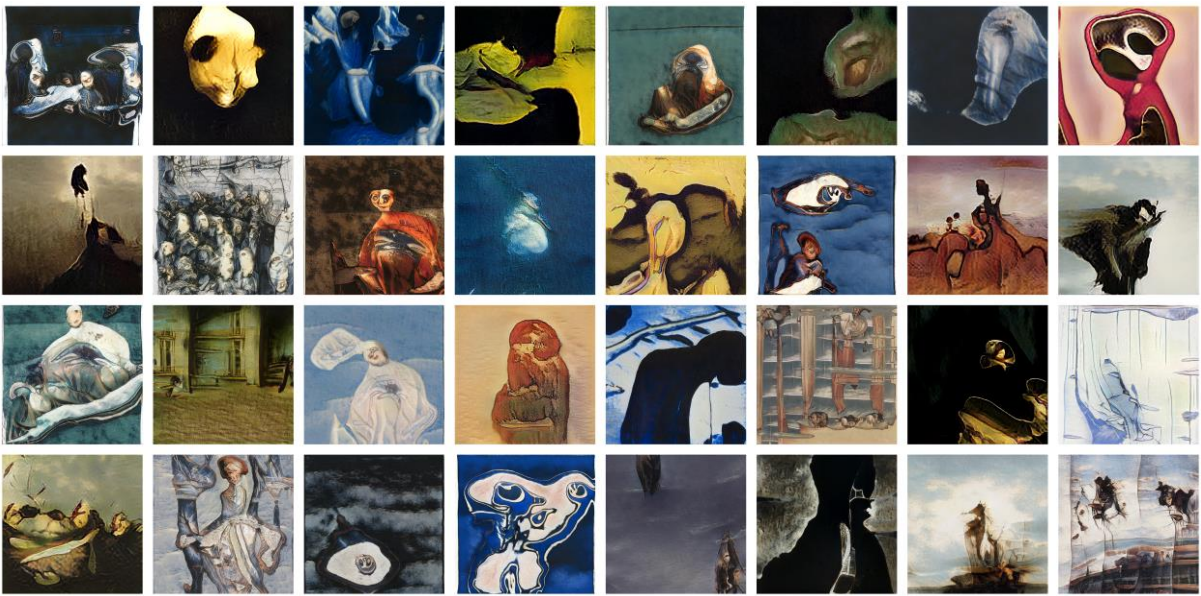

*Figure 2: Paintings belonging to the class "sadness".*

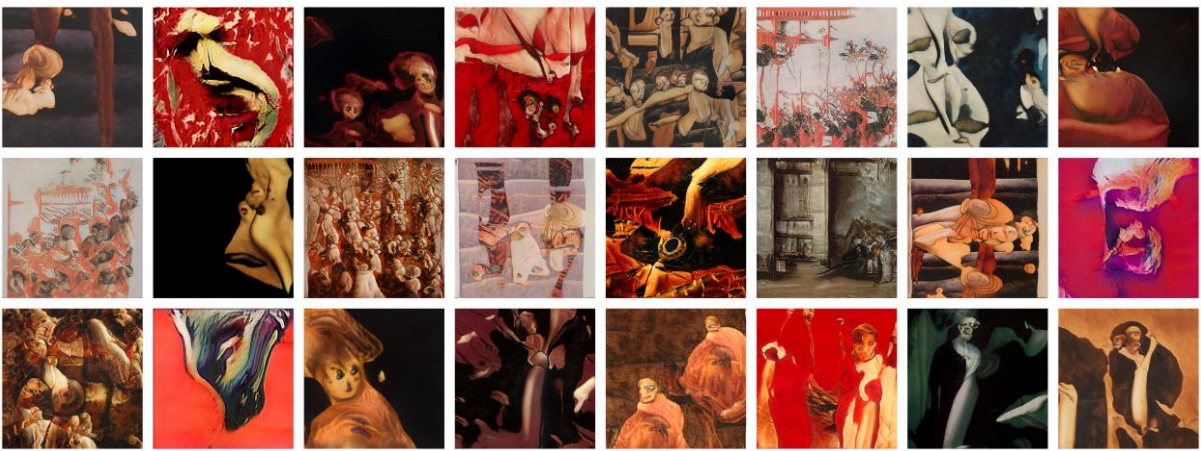

*Figure 3: Paintings belonging to the class "anger".*

In "fear" (Figure 4), some of the paintings represent threatening characters immersed in bleak scenarios, and it seems that their goal is to strike fear in the spectator. The remaining ones depict fear from the point of view of who feels it. This representative dualism seems to make these paintings notably heterogeneous in both colour palettes and shapes. Paintings in "happiness" (Figure 5) stand out because of their bright and gaudy colours. Most of them represent bucolic scenarios with green grass and blue skies. Different from the previous classes, the existing characters transmit a sense of joy and tranquillity.

The expressivity and shape variability of the paintings make this system emerge compared to its related works. Several of the proposed paintings do not precisely fall into the idea of their assigned emotion, or they evoke more emotions at once. This effect is not unexpected or undesired, as we have made explicit design choices (see Section 2) to obtain it.

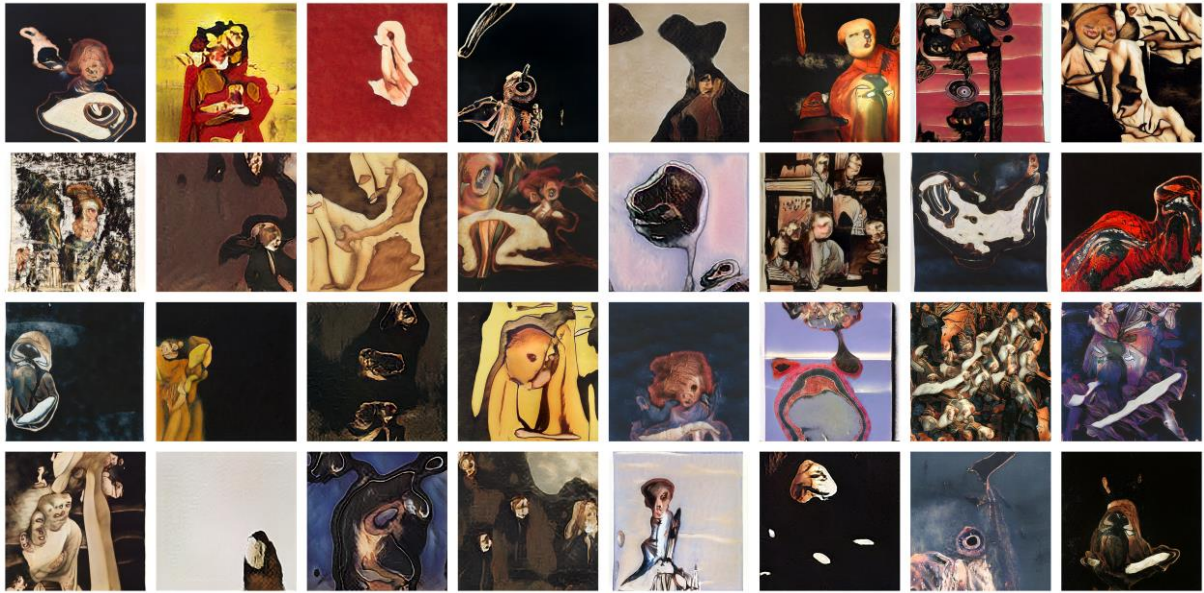

*Figure 4: Paintings belonging to the class "fear".*

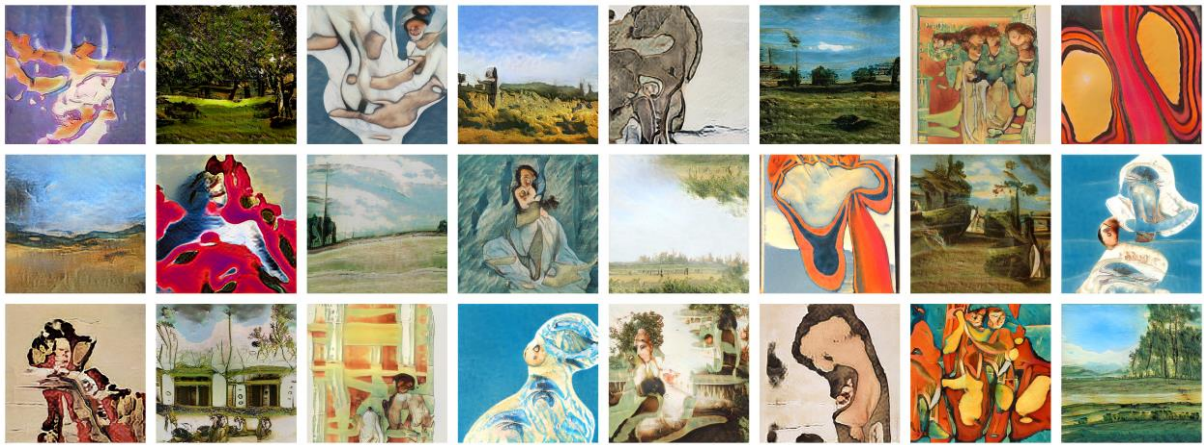

*Figure 5: Paintings belonging to the class "happiness".*

## 5. ART THERAPY AND DISABILITY AESTHETICS

Critical Disability Studies is an active and interdisciplinary research area characterized by several debates and enigmas (Watson and Vehmas, 2019) in the analysis of disability as a "cultural, historical, relative, social, and political phenomenon" (Hall, 2019), with an activist component. Given the nature of our work, we analyse our system from the perspective of the intersection between art and disabilities, focusing on art therapy and disability aesthetics.

With the term "art therapy" we refer to art practices that can improve the well-being of patients (Solvang, 2018). In our system, we provide the possibility of creating paintings from the brain activity during an emotional experience, suggesting that artistic abilities are not necessarily attached to physical bodies. Despite the lack of experiments that can confirm this, we believe that the utilization of such a system could improve self-esteem and marginalization issues for some individuals, and we suggest that further studies on this topic would be highly beneficial.

The term "disability aesthetics" refers to the historical connection between the conception of the beautiful in art and the stigmatisation of disabilities (Levin and Siebers, 2010). People with disabilities have been marginalised from the art discourse on two different levels: on the one hand, disabilities have been highly under-represented in the history of art, and, on the other hand, the art narrative has silenced the stories and intentions of artists with disabilities (Sherwood, 2019). The first aspect of this marginalisation has seen a turn-out in modern art, establishing new aesthetic norms that allow a more diverse representation of humanity. This sensibility is evident in the techniques of Dada and Expressionism, which rely on deforming bodies and expressing high artistic ideas through "imperfections". According to the theories of Tobin Siebers (Siebers, 2005), modern art has allowed the representation of disability to become an aesthetic value.

Linking our results to this discourse, we raise questions regarding the role of AI generative art in establishing new aesthetic norms or in reinforcing existing ones. Our AI system is trained on paintings representing human bodies with clear, defined, and symmetric shapes. However, the scarcity of training data and their heterogeneity imply that the system cannot reproduce human figures perfectly. Despite this technical limit, the resulting paintings can express the emotions in the inputted EEG signals, demonstrating that imperfections can be at the base of emotional expression in art practices. The shapes in our paintings suggest that AI can be "freer" than humans from prejudices on the canonical ideas of beauty and, in such a context, we wonder whether the progressive acceptance of AI-generated paintings as part of the "high art" could lead to an increased aesthetic acceptance of a more diverse and inclusive representation of the human condition.

## 6. THEORETICAL DISCUSSION

The BCI described in this paper is related to two technical research areas: affective computing and computational creativity. Affective computing studies the interaction between humans and computers when computers can communicate on an emotional or affective level (Picard, 1999). Computational creativity, instead, investigates the relationship between computational systems, creative practices, and the concept of creativity (McCormack and d'Inverno, 2014). Being a BCI that interprets human emotions in EEG signals and translates the emotions into paintings, our work is an intersection between these two areas.

One of the remarkable features of the proposed system is that the user is passive in the final emotional expression. The lack of control over the generation implies that the machine receives trust in its affective and artistic abilities. We are, therefore, proposing a software that blurs the anthropological border between humans and machines, allowing a different conceptualization of this interaction (Suchman and Weber, 2016) as the two entities contribute **as equals** to reach an artistic outcome.

We conceive this BCI as a creative extension of human bodies. In *The medium is the message* (McLuhan and Fiore, 1967), McLuhan suggests that every extension of ourselves is a medium and that media carry "messages", intrinsic in the way they modify the pace and scale of our actions in society. In our work, it is fascinating to note that the discussed technology brings novel artistic opportunities whose "message" is still to be understood.

If we accept the idea of this BCI expressing emotions through paintings, then it is crucial to investigate the norms that it adopts for such emotional expression, considering art as an evolving phenomenon reflecting surrounding values. Despite the strong connection between the generated images and the training ones, it is still possible to imagine that this system leverages different paradigms than human artists. As such, the system obtains a "power" that overcomes the intended will of the human providing the EEG signal. Relating this work to classical philosophical questions and with Kant's intersubjectivity of the aesthetic taste, we wonder whether some of the technical adjustments described in Section 3.2 could contribute to simulating an artistic expression based on a greater logic, truth, or rationality that is outside and above the individual human being.

## 7. CONCLUSION

We have presented and discussed an AI-based BCI that translates emotional states from EEG signals into original paintings, implementing a humanised software that "understands" emotions and expresses them artistically. We have discussed future work directions to overcome technical and ethical limitations of this system, which, on the other hand, shows high potentialities in the fields of art therapy and disability aesthetics. Our results and discussion contribute to inquiring the role of machines in our complex cultural gear from a classical and anthropological perspective.

## 8. REFERENCES

Bergaust, K. & Nichele, S. *FeLT-The Futures of Living Technologies.* DOI: 10.14236/ewic/POM19.14